\documentclass[runningheads]{llncs}
\usepackage{graphicx}
\usepackage{colortbl}
\definecolor{Gray}{gray}{0.9}
\usepackage{url}
\usepackage{cite}
\usepackage{ragged2e}
\justifying

\begin{document}
\title{Exploring the Transferability of a Foundation Model for Fundus Images: Application to Hypertensive Retinopathy}
\titlerunning{Exploring the Transferability of a Foundation Model for Fundus Images}

\author{
Julio Silva-Rodriguez\inst{1} \and
Jihed Chelbi\inst{2} \and
Waziha Kabir\inst{2} \and
Hadi Chakor\inst{2} \and
Jose Dolz \inst{1} \and
Ismail Ben Ayed \inst{1} \and
Riadh Kobbi\inst{2}
}

\authorrunning{Silva-Rodriguez et al.}
\institute{
ETS Montreal, Quebec, Canada \\
\email{julio-jose.silva-rodriguez@etsmtl.ca} \and
DIAGNOS Inc., Quebec, Canada}

\maketitle

\begin{abstract}

Using deep learning models pre-trained on Imagenet is the traditional solution for medical image classification to deal with data scarcity. Nevertheless, relevant literature supports that this strategy may offer limited gains due to the high dissimilarity between domains. Currently, the paradigm of adapting domain-specialized foundation models is proving to be a promising alternative. However, how to perform such knowledge transfer, and the benefits and limitations it presents, are under study. The CGI-HRDC challenge for Hypertensive Retinopathy diagnosis on fundus images introduces an appealing opportunity to evaluate the transferability of a recently released vision-language foundation model of the retina, FLAIR \cite{FLAIR}. In this work, we explore the potential of using FLAIR features as starting point for fundus image classification, and we compare its performance with regard to Imagenet initialization on two popular transfer learning methods: Linear Probing (LP) and Fine-Tuning (FP). Our empirical observations suggest that, in any case, the use of the traditional strategy provides performance gains. In contrast, direct transferability from FLAIR model allows gains of $\sim2.5\%$. When fine-tuning the whole network, the performance gap increases up to $\sim4\%$. In this case, we show that avoiding feature deterioration via LP initialization of the classifier allows the best re-use of the rich pre-trained features. Although direct transferability using LP still offers limited performance, we believe that foundation models such as FLAIR will drive the evolution of deep-learning-based fundus image analysis.

\keywords{Foundation Models \and Transfer Learning \and Hypertensive \\ Retinopathy.}
\end{abstract}

\section{Introduction}
\label{sec:1}

\begin{figure*}[ht!]
\begin{center}

\includegraphics[width=1\textwidth]{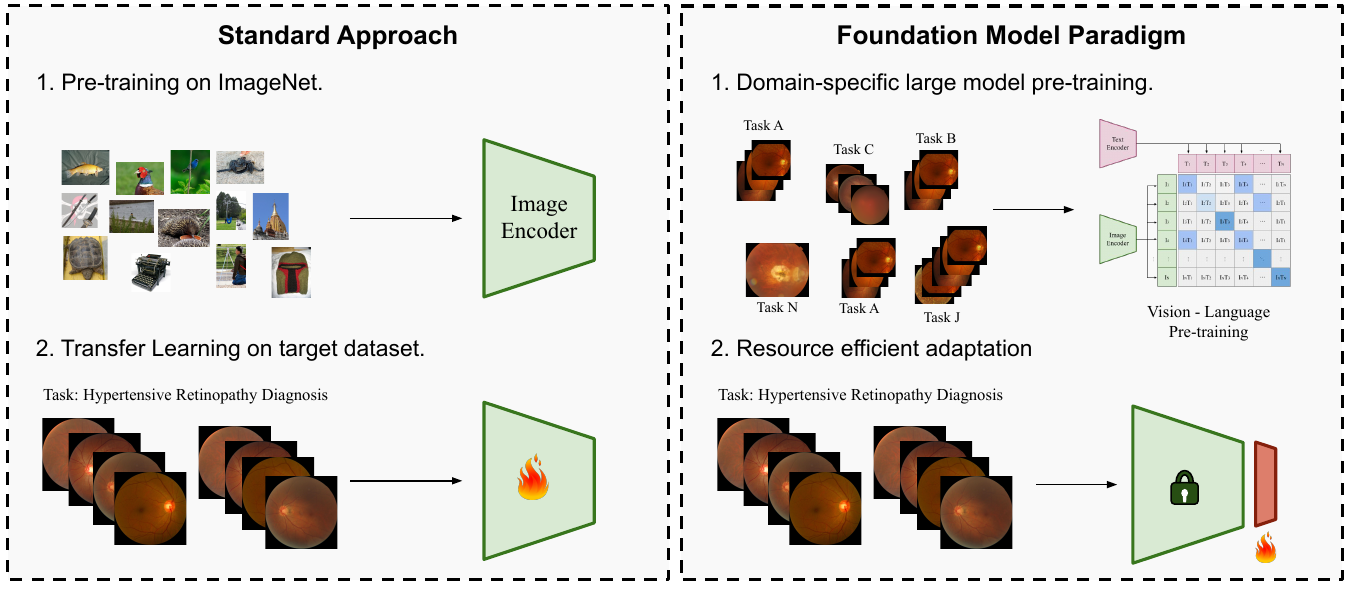}

\caption{\textbf{Standard vs. Foundation Model Paradigms}. Deep learning solutions on medical image analysis are traditionally built upon models pre-trained on ImageNet to alleviate the need for large datasets. Nevertheless, the benefits of transfer learning might be limited when a substantial domain gap from source to target exists \cite{raghu2019transfusion}. Foundation models on specific domains, such as FLAIR \cite{FLAIR} for fundus image analysis, which is pre-trained on heterogeneous data sources and tasks, offer better resource-efficient transferability to new tasks.}
\label{fig:summary}
\end{center}
\end{figure*}

A foundation model for image understanding is a generic pre-trained deep learning model on a large dataset, serving as a base for developing specialized vision models through fine-tuning on task-specific data. Recently, foundation models trained on natural images have gained popularity by the impressive resource-efficient transferability capabilities they present. Successful examples include pre-trained models on ImageNet, vision-language pre-training as CLIP \cite{Radford2021} or ALIGN \cite{jia2021scaling}, or models for image segmentation as SAM \cite{kirillov2023segany}. Despite its promising results in the natural image context, these models have shown limited performance for transferability to expert fields such as medical image analysis \cite{Wang2022, SAM_Problem_1, SAM_Problem_2}. Although the limited benefit of using transfer learning from large pre-trained models when exists a large domain gap is not new \cite{raghu2019transfusion}, these observations have encouraged the recent development of foundation models specialized in concrete medical domains (see Figure \ref{fig:summary}). As a result, a paradigm shift is occurring in this field. The use of specialized foundation models promises to improve the efficiency of the resources needed to create task-specific solutions, in both samples and computational power. Some successful models have been developed for radiology \cite{Wang2022}, histology \cite{HistoFound}, fundus images \cite{FLAIR}, volumetric segmentation \cite{Liu2023CLIPDriven, silvaRodriguez2023fseft}, and 2D image segmentation \cite{Butoi2023}. However, the potential of the \textit{pre-train and adapt} paradigm remains largely unexplored in many medical imaging domains. This motivates the realization of empirical studies to analyze the benefits of such models in comparison with the more traditional paradigms.

The CGI-HRDC Challenge for Hypertensive Retinopathy diagnosis through fundus images constitutes an ideal setting to study the potential of foundation models. The analysis of hypertensive retinopathy is burdened by the necessary manual inspection of fundus images from experienced ophthalmologists. Therefore, it is paramount to provide ophthalmologists with an accurate computer system that facilitates the analysis of the course of the disease. Moreover, the scarcity of available data sources with hypertensive cases further challenges the development of task-specific deep learning models. Thus, the objective of this work is to study the limitations and potential of a recently released foundation model for fundus image analysis, FLAIR \cite{FLAIR}, and compare its transferability capabilities for Hypertensive Retinopathy detection, in comparison with standard solutions using models pre-trained on Imagenet. 

\section{Related Works}
\label{sec:2}

\subsection{Transfer learning on fundus images}

Deep learning has achieved remarkable performance on a wide variety of fundus image analysis tasks, and offers a potential solution for large-scale screening and early detection of ophthalmologic conditions \cite{Balyen2019, Bellemo2019}. Among others, outstanding applications include diabetic retinopathy grading \cite{ChandrasekaranRetinopathy2023, Liu2022}, cataract diagnosis \cite{ImranFundus2020}, lupus detection \cite{LiuTMM2020}, or multi-disease classification \cite{AbdulRDD2022, Jin2022}. Nevertheless, training such models from scratch demands substantial datasets and extensive computational resources \cite{Erhan2009}. In the medical domain, specifically in fundus image analysis, achieving the prerequisite of large datasets is often unattainable, and the norm involves working with small, task-specific datasets. Consequently, transfer learning from natural images has emerged as the primary approach for medical image classification \cite{raghu2019transfusion}. However, empirical studies have revealed that transfer learning may yield limited performance improvements in specific medical image classification scenarios \cite{raghu2019transfusion, Neyshabur2020}, in which a large inter-domain gap exists \cite{Azizpour2014}. These limitations have motivated the use of pre-trained models for further transferability to downstream tasks. For example, self-supervised \cite{Srinivasan2022} or task-specific pre-training \cite{Liu2022} using public datasets have shown promising improvements for diabetic retinopathy grading. 
However, it is important to note that task-specific models are prone to produce too specific inductive biases on specific features, resulting in poor generalization when transferred to other less-related tasks \cite{FLAIR}. In this context, vision-language pre-training has raised as a promising solution to group heterogeneous data sources and tasks for pre-training, aligned through text supervision, and thus capturing generic features and representations in large foundation models. This strategy has shown promising transferability performance in the medical context for radiology \cite{Wang2022}, histology \cite{HistoFound}, and recently in fundus images \cite{FLAIR}.

\subsection{FLAIR}

The foundation model FLAIR\footnote{\scriptsize Available at \url{https://github.com/jusiro/FLAIR}} \cite{FLAIR} (A Foundation LAnguage Image model of the Retina) is a recently released pre-trained model for universal disease detection on fundus images through text supervision, which has shown remarkable transferability to downstream tasks even on unseen diseases.

\noindent\textbf{\textit{FLAIR pre-training datasets}}. The foundation model was built using an assembly dataset from $37$ publicly available sources, which include up to 286,916 fundus images from heterogeneous tasks, consisting of $96$ different categories. These tasks include \textit{diabetic retinopathy grading}: EYEPACS\footnote{\scriptsize\url{https://www.kaggle.com/c/diabetic-retinopathy-detection}}, IDRID \cite{Porwal2020}, JICHI \cite{Takahashi2017}, PARAGUAY \cite{PARAGUAY}, SYSU \cite{SYSU}, OIA-DDR \cite{DDR} and BRSET \cite{BRSET}; \textit{Glaucoma detection}: LAG \cite{LAG_Li2019} , PAPILA \cite{Kovalyk2022}, CHAKSU \cite{chaksu} and AIROGS (\cite{AIROGS}); \textit{lesion segmentation}: DR1-2 \cite{Pires2014}, SYSU \cite{SYSU}, OIA-DDR \cite{DDR} and HEI-MED \cite{Giancardo2012}; \textit{image description}: EYENET \cite{Huang2021}, ODIR-5K\footnote{\scriptsize\url{https://odir2019.grand-challenge.org/}}, and STARE \cite{Hoover2000, Hoover2003}; and the detection of \textit{other diseases}: RFMid \cite{Pachade2021}, 1000x39 \cite{1000x39}, BRSET \cite{BRSET} and FUND-OCT1 \cite{FUND-OCT1, FUND-OCT2}. From the last group, it is worth mentioning that nearly $400$ samples from two different datasets contained hypertensive retinopathy findings, which constitutes less than $0.2\%$ of the entire assembly dataset.

\noindent\textbf{\textit{Model architecture}}. FLAIR model consists of a vision encoder, ResNet-50 \cite{He2016}, and a text encoder, with the architecture of BioClinicalBert\footnote{\scriptsize\url{https://huggingface.co/emilyalsentzer/Bio_ClinicalBERT}}, which takes as input a fundus image and a text prompt describing its content, respectively. The produced individual modality embeddings are projected into an l2-normalized multimodal space. 

\noindent\textbf{\textit{Optimization criteria}}. The foundation model is pre-trained using a contrastive vision-language alignment approach, aiming to create a multimodal feature representation in which images and expert knowledge descriptors of the same category are similar while maximizing differences between unrelated samples. This three-dimensional alignment, encompassing image, text, and categories, results in a more comprehensive and richer representation through text semantics, able to inter-correlate different conditions (e.g. diabetic retinopathy and microaneurysms) by efficiently leveraging expert domain knowledge.

\subsection{Transferability}

In the context of foundation models, transferability refers to the process of using or adapting the features learned in large pre-trained models to downstream tasks and related domains. In this work, we focus on the transferability in the medium data regime, where a few hundred training examples are available, and we explore only adaptation through the vision encoder. Two popular transfer learning methods are Linear Probing (LP) and Fine-Tuning (FT). The former involves direct transferability of the features by adjusting only the linear classifier. For the latter, all the parameters of the model are re-trained to the target dataset. Fine-tuning all layers of a network can modify the pre-trained features by adapting/improving them to the downstream task, while linear probing, on the other hand, only relies on the frozen features without any further adjustments.

\section{Method: Transfer Learning from FLAIR model}
\label{sec:3}

In this work, we aim to explore the potential and limitations of transferring a general-purpose foundation model of the retina for the challenging task of Hypertensive Retinopathy. In particular, we focus on adapting the image encoder from the recently published FLAIR \cite{FLAIR} model.

\noindent\textbf{\textit{Pre-processing.}} The fundus images are processed accordingly to the foundation model pre-training. Concretely, the samples are resized to $800\times800$ pixels, and the intensity is scaled between $[0, 1]$.

\noindent\textbf{\textit{Linear Probe (LP) adaptation.}} For LP adaptation, a classification head is trained over the features extracted from the pre-trained FLAIR model. Two feature representations are considered for LP adaptation: the vision encoder representation (LP (vision)), and the multimodal vision-language projection (LP (proj)).

\noindent\textbf{\textit{Fine-Tuning (FT).}} In this setting, a classification head is initialized with random weights, which uses as input the vision encoder features, and the whole network is retrained on the target task. Concretely, the encoder and classifier are trained to minimize the binary cross-entropy between reference and predicted sigmoid scores via stochastic gradient descent.

\noindent\textbf{\textit{Linear Probe and Fine-Tuning (LP+FT).}} Last, we follow a recently popularized two-step strategy. First, the classifier is trained with the backbone frozen as in LP, and then the whole network is regularly fine-tuned to the objective task \cite{Kumar22, Kanavati2021}.

\section{Experiments}
\label{sec:4}

\subsection{Dataset}

The CGI-HRDC dataset comprises two different tasks: Task 1 involves hypertension classification, determining whether the patient has hypertension, while Task 2 focuses on Hypertensive Retinopathy detection, aiming to identify signs of Hypertensive Retinopathy in the target fundus image. For each task, the development dataset includes 712 samples for training. In addition, the Challenge includes 288 cases for testing for each task, which remain unavailable during the development stage. The samples consist of macula-centered fundus images, each with dimensions of $800\times800$ pixels.

\subsection{Implementation details} The pre-trained FLAIR vision encoder is transferred to the different tasks related to hypertensive retinopathy diagnosis using the strategies indicated in Section \ref{sec:3}. For LP adaptation, We follow the same solver as in CLIP \cite{Radford2021}, and we applied class weights to account for class imbalances. For full backbone fine-tuning, ADAM is used as an optimizer with an initial learning rate of $1e-4$, and training is carried out using mini-batches of $4$ images, during $20$ epochs. To account for class imbalance, a re-sampling strategy of the minority class is carried out. Data augmentation is applied for each iteration using random horizontal flips, rotations of $[-5, 5]$ degrees, zoom scaling in the range $[0.9, 1.1]$, and color jitter. Also, the convergence is tracked on the internal validation set, and the best model in this subset is saved as the final solution for evaluation. For each stage of LP+FT method, we follow the same aforementioned implementation details. The adaptation code was part of the official FLAIR repository, publicly accessible at: \small{\url{https://github.com/jusiro/FLAIR}}.

\subsection{Baselines} To evaluate the benefits of using a domain-specific foundation model for transferring feature representations, we use the ResNet-50 \cite{He2016} (the same vision backbone used in FLAIR) with weights pre-trained on ImageNet \cite{Deng2009}, for natural image classification. In particular, the different transfer learning strategies set for FLAIR are applied to this model for adaptation to the challenge tasks. The hyperparameters and implementation details of these baselines were the same as the foundation model adaptation. Hereafter, we refer to this weights initialization as \textit{Imagenet}.

\subsection{Evaluation protocol and metrics}

During the method development stage, a $5$ fold cross-validation partition is performed on the CGI-HRDC development dataset to evaluate the different proposed methods. In each fold iteration, $20\%$ of training samples for each class are randomly retrieved for evaluation, while $70\%$, is used for training and $10\%$ for internal validation. The evaluation metrics used are the Kappa, F1 score, and specificity, which are averaged into a global score. All metrics are averaged fold-wise during the cross-validation stage.

\section{Results}
\label{sec:5}

\subsection{Development dataset results}

The cross-validation results obtained in the training subset using the different strategies for adapting FLAIR model and the corresponding baselines for hypertensive classification (Task 1) and Hypertensive Retinopathy detection (Task 2) are presented in Table \ref{table_task1} and Table \ref{table_task2}, respectively.
\begin{table*}[h!]
\centering
\caption{\textbf{Cross-Validation results for Task 1: Hypertensive classification} LP: Linear Probe; FT: Fine-Tuning; proj: projection. Gray indicates the method submitted for the testing phase.}
\label{table_task1}
\begin{tabular}{lcccc}
\hline
                           & \multicolumn{4}{c}{Metric}          \\ \cline{2-5}
\multicolumn{1}{c}{Method} & Kappa & F1    & Specificity & Avg.  \\
\hline
\textit{Imagenet} - LP              & 0.324(0.039) & 0.666(0.019)  & 0.651(0.035)       & 0.547 \\
\textit{Imagenet} - FT              & 0.335(0.112) & 0.659(0.078)  & 0.682(0.019)       & 0.558 \\
\textit{Imagenet} - LP+FT           & 0.389(0.074) & \textbf{0.711}(0.023)  & 0.637(0.113)       & 0.579 \\ \hline
FLAIR - LP (proj)                   & 0.240(0.037) & 0.593(0.017)  & 0.685(0.051)       & 0.506 \\
\cellcolor{Gray}FLAIR - LP (vision) & \cellcolor{Gray}0.358(0.066) & \cellcolor{Gray}0.680(0.033) & \cellcolor{Gray}0.676(0.035)       & \cellcolor{Gray}0.571 \\
FLAIR - FT                          & 0.366(0.110) & 0.697(0.039)  & 0.640(0.121)       & 0.567 \\ 
FLAIR - LP+FT                       & \textbf{0.420}(0.043) & 0.703(0.026)  & \textbf{0.730}(0.058)       & \textbf{0.617} \\ \hline
\end{tabular}
\end{table*}

\begin{table*}[h!]
\centering
\caption{\textbf{Cross-Validation results for Task 2: Hypertensive Retinopathy classification} LP: Linear Probe; FT: Fine-Tuning; proj: projection. Gray indicates the method submitted for the testing phase.}
\label{table_task2}
\begin{tabular}{lcccc}
\hline
                           & \multicolumn{4}{c}{Metric}          \\ \cline{2-5}
\multicolumn{1}{c}{Method} & Kappa & F1    & Specificity & Avg.  \\
\hline
\textit{Imagenet} - LP              & 0.404(0.068) & 0.652(0.040)  & 0.740(0.040)       & 0.598 \\
\textit{Imagenet} - FT              & 0.623(0.049) & 0.770(0.030)  & 0.874(0.049)       & 0.755 \\
\textit{Imagenet} - LP+FT           & 0.636(0.103) & 0.781(0.061)  & 0.869(0.049)       & 0.762 \\ \hline
FLAIR - LP (proj)                   & 0.258(0.089) & 0.533(0.068)  & 0.759(0.045)       & 0.516 \\
\cellcolor{Gray}FLAIR - LP (vision) & \cellcolor{Gray}0.439(0.052) & \cellcolor{Gray}0.670(0.033) & \cellcolor{Gray}0.764(0.034)       & \cellcolor{Gray}0.624 \\
FLAIR - FT                          & 0.622(0.027) & 0.772(0.017)  & 0.862(0.062)       & 0.752 \\ 
FLAIR - LP+FT                       & \textbf{0.695}(0.060) & \textbf{0.816}(0.034)  & \textbf{0.893}(0.062)       & \textbf{0.801} \\ \hline
\end{tabular}
\end{table*}

The obtained results unveil the benefit of using foundation models pre-trained on medical domains. \textbf{\textit{Linear Probe (LP) adaptation.}} Direct transferability (i.e. LP) - of FLAIR features improves in $\sim+2.5\%$ the score compared to \textit{Imagenet} features on both Tasks. It is worth mentioning that, in the case of FLAIR, using the features of the multimodal projection results in a significant performance drop. Despite this feature representation is commonly used for the transferability of vision-language pre-trained models on other works (e.g. CLIP \cite{Radford2021}, MedCLIP \cite{Wang2022}), our empirical results evidence that they might produce suboptimal solutions. This may be caused by the specific patterns of Hypertensive Retinopathy, and the low prevalence of this condition in the FLAIR pre-training dataset ($<0.2\%$). Thus, tuning the vision encoder for this task seems necessary in this case. \textbf{\textit{Fine-Tuning (FT).}} After fine-tuning, the obtained performance increases notably for Task 2, while modest improvements are observed for Task 1. In this case, minor differences between \textit{Imagenet} and FLAIR initialization can be observed. Interestingly, in the case of Task 1, just LP outperforms FT for the whole network. As it is widely known, full FT is an aggressive adaptation strategy, which might distort pre-trained features \cite{Kumar22}. \noindent\textbf{\textit{Linear Probe and Fine-Tuning (LP+FT).}} When using the classifier initialized via LP, then the use of a domain-specific Foundation model highlights its benefits. This solution prevents the distortion of pre-trained features, and the performance consistently improves in $\sim+4\%$ compared to using \textit{Imagenet} representations. Although the benefits of LP+FT observations have been previously reported for regular fine-tuning \cite{Kanavati2021} and out-of-distribution inference \cite{Kumar22}, our empirical results suggest that the quality of the initialization features and classifier for the target domain also plays an important role in this setting. \textbf{\textit{Performance discrepancies between tasks.}} The results obtained in Task 1 are consistently worse compared to the performance of the models observed in Task 2. This might be produced by the hardness of the target case. While Hypertension might be a global condition of the patient, with scarce feature representation on the particular eye of the sample, Hypertensive Retinopathy ensures the presence of a disease in the retina of the target fundus image.

\subsection{CGI-HRDC hidden test results}

After the development stage, we decided to use the Linear Probe adaptation with the FLAIR vision encoder features (i.e. FLAIR - LP (vision) in Tables \ref{table_task1} and \ref{table_task2}) as our solution for the CGI-HRDC challenge. Although this was not the best method in the cross-validation set, the motivation behind this decision was to test the direct transferability of the foundation model in a real use case. Thus, a classifier for each task was trained on top of the frozen vision encoder of FLAIR using the whole challenge development dataset. Under this setting, a global average score of $0.500$ ($\#3rd$ on the official test Leaderboard) and $0.545$ ($\#2nd$ on the official test Leaderboard) was obtained for Task 1 and Task 2, respectively. It is worth mentioning that the proposed method experiences a consistent drop of $\sim-8\%$ with respect to the cross-validation stage which might be caused by disparities in class balance or the presence of harder samples on the hidden test subset. 

\section{Conclusions}
\label{sec:6}

In this work, we have explored the transferability of a foundation model for fundus images, FLAIR \cite{FLAIR}, to tasks related to Hypertensive Retinopathy detection, in the context of the CGI-HRDC challenge. FLAIR model, although pre-trained through contrastive vision-language alignment in a wide variety of Fundus conditions, contains less than $0.2\%$ of training samples with pathologies related to hypertension. Still, the learned feature representations show promising capability for direct transferability on such a challenging task, with gains of $\sim+4\%$ compared to pre-training on \textit{Imagenet}. Nevertheless, the modest results obtained using Linear Probing in comparison with other methods participating in the challenge highlight the current limitations of direct transferability for reaching state-of-the-art performance in medium-sized datasets. Thus, we have explored fine-tuning the whole model for adaptation. In any case, using the model pre-trained on \textit{Imagenet} - which is the \textit{de-facto} solution on transfer learning for medical image analysis - has shown any advantage compared to using FLAIR. In particular, preventing feature distortion of the Foundation model through Linear Probing initialization showed promising benefits for both tasks. We believe that developing foundation models on medical domains and enhancing the adaptation of their rich feature representations to downstream tasks is an appealing future direction for medical image analysis and, more specifically, for the characterization of fundus images. 

\section*{Acknowledgments}

The work of J. Silva-Rodríguez was partially funded by the \textit{Fonds de recherche du Québec (FRQ)} under the Postdoctoral Merit Scholarship for Foreign Students (PBEEE).

%
%
%
\bibliographystyle{splncs04}
\bibliography{refs}

\end{document}